\documentclass[conference]{IEEEtran}
\usepackage{times}
\usepackage{latexsym}
\usepackage{amsmath}
\usepackage{amssymb}
\usepackage{multirow}
\usepackage{graphicx}

\title{A retrieval-based dialogue system utilizing utterance and context embeddings}

\author{\IEEEauthorblockN{Alexander Bartl and Gerasimos Spanakis}
\IEEEauthorblockA{Department of Data Science and Knowledge Engineering\\
Maastricht University\\
6200MD, Netherlands\\
Email: a.bartl@student.maastrichtuniversity.nl,jerry.spanakis@maastrichtuniversity.nl}}

\begin{document}

\maketitle

\footnote{A version of this paper is accepted at ICMLA2017 conference http://www.icmla-conference.org/icmla17/}

\begin{abstract}
Finding semantically rich and computer-understandable representations for textual dialogues, utterances and words is crucial for dialogue systems (or conversational agents), as their performance mostly depends on understanding the context of conversations. Recent research aims at finding distributed vector representations (embeddings) for words, such that semantically similar words are relatively close within the vector-space. Encoding the ``meaning" of text into vectors is a current trend, and text can range from words, phrases and documents to actual human-to-human conversations. In recent research approaches, responses have been generated utilizing a decoder architecture, given the vector representation of the current conversation. In this paper, the utilization of embeddings for answer retrieval is explored by using Locality-Sensitive Hashing Forest (LSH Forest), an Approximate Nearest Neighbor (ANN) model, to find similar conversations in a corpus and rank possible candidates. Experimental results on the well-known Ubuntu Corpus (in English) and a customer service chat dataset (in Dutch) show that, in combination with a candidate selection method, retrieval-based approaches outperform generative ones and reveal promising future research directions towards the usability of such a system.
\end{abstract}


\section{Introduction}
Text-only based Dialogue systems, also called Conversational Agents, Chatbots or Chatterbots, have become very popular in the research community and for large companies. The reason for the rise in popularity lies in the fact that their ability to interact intelligently with humans has improved significantly due to advancements in hardware technologies and Artificial Intelligence.

One of the latest effective approaches \cite{mikolov2013distributed} is to represent words, phrases, or even complete dialogues as fixed-length vectors of floating point numbers, also called embeddings (or distributed representations or feature vectors). The Hierarchical Recurrent Encoder-Decoder (HRED) \cite{serban2016building} and its successors \cite{serban2016hierarchical}, (as well as similar related models) are specifically designed to encode the meaning of textual conversations regarding the special structure that originates from multiple turn-taking speakers. 

In our approach, a context embedding, a vector encoding the meaning of a conversation up to a certain time step $t$, encoded by the HRED model, serves as input to the decoder component to generate a textual answer. We explore the performance of a retrieval-based model that uses the utterance- and context-embeddings, previously generated by the HRED model, to find similar conversations and rank possible candidate answers. We argue that a retrieval-based approach, based on embeddings, can outperform the generative approach, as the retrieval of similar conversations is less dependent on high quality embeddings and less susceptible to poorly trained embeddings.

The rest of the paper is organized as follows: We first give an outline of the research around dialogue systems, showing the development from purely script-based systems to deep networks generating answers end-to-end. The proposed pipeline is discussed in Section \ref{model_description}. Experimental setup (datasets, evaluation metrics, models implemented and compared) as well as the results are discussed in Section \ref{experiments}. Finally, we conclude the paper by summarizing the main findings and outline future work. 

\section{Related Work} \label{related_work}

The purpose of Dialogue Systems (DS), often also termed Conversational Agents (CA) or Chatterbots, is to converse with humans to provide information, help in decision making, perform administrative services, or just for the sake of entertainment \cite{shawar2007chatbots}. The traditional design of Dialogue Systems \cite{jurafsky2000speech} follows a modular approach, splitting the system usually into a \textit{Natural Language Understanding} (NLU) module, a \textit{Dialogue Manager} and a \textit{Natural Language Generation} (NLG) unit. The NLU module processes the raw user input and extracts useful information and features that can be used by the Dialogue Manager to update internal states, send queries to a knowledge base or, more generally, find actions based on a script. The NLG acts inversely to the NLU module, receiving features and information from the Dialogue Manager to generate a response that, finally, will be presented to the user.


One of the simplest design approaches for an NLU is to simply spot certain key-words or combinations of them. This is often the general procedure of script-based chatbots and the approach followed by ELIZA. However, there is a long history of attempts \cite{bates1995models} to improve NLU and find better representations of text. With advances in machine learning, the development ranges from statistical modelling of language \cite{manning1999foundations}, semantic parsing \cite{dowding1993gemini}, skip-gram models \cite{mikolov2013distributed}, and others, to approaches utilizing deep neural architectures \cite{collobert2008unified,lecun2015deep}. Neural networks have also been used to improve NLG \cite{vinyals2015neural, sutskever2014sequence}. Recently, Dialogue Managers have made similar advances towards automated solutions, with a focus on reinforcement learning \cite{shah2016interactive, young2013pomdp}, generating policies of how to interact with humans, based on some state representation.

With the rise of Deep Learning (DL) in recent years \cite{schmidhuber2015deep} and an increasing company interest in chatterbots, end-to-end Dialogue Systems, such as deep Recurrent Neural Networks, constituting all modules in one model \cite{serban2016building}, have become one of the major research topics for Dialogue Systems. The tasks of NLU, NLG, and the Dialogue Manager are performed by a single deep network that is trained to reproduce conversations from a large dataset. Such a system would generate an answer end-to-end from raw user input. Even though training deep RNNs can be considerably difficult \cite{bengio2013advances}, only one model would need to be optimized, and one could benefit from the neural model's capability to generate natural responses \cite{wen2015stochastic, vinyals2015neural}. 

Our proposed pipeline is a combination of a generative- and retrieval-based approach. An encoder model, such as the HRED model (or could be one of its more advanced variations) is trained end-to-end on a textual corpus, using an objective function that is based on how capable the model is of generating the answers in the training set. After the training however, the decoder component of the HRED model is not used to generate answers. Instead, we argue that a retrieval-based approach taking over the NLG part performs equally or better in both general and specific domains.

\section{Model description} \label{model_description}
\begin{figure*}[h]
\centering
\scalebox{1}{
\includegraphics[width=1\textwidth]{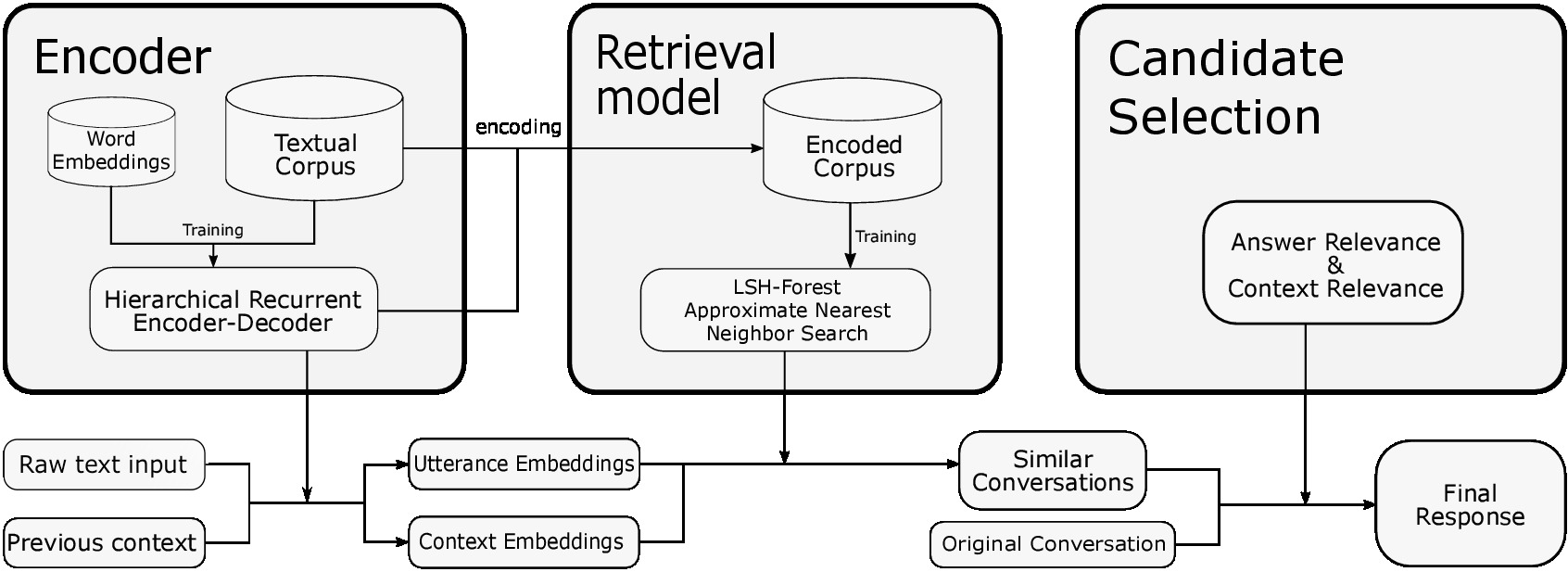}}
\caption[A view of the pipeline implementing the proposed approach]{A view of the pipeline implementing the proposed approach. An HRED encoder is used to generate context and response embeddings and an ANN model builds on previous steps to retrieve similar conversations. Finally, the best candidate is selected according to answer- and context-relevance.}
    \label{fig:pipelineoverview}
\end{figure*}

The proposed model can be split up into three individual components. The first component, the encoder, utilizes the HRED model to encode raw conversations into embeddings containing the actual meaning. The second component, a retrieval-based approach using an Approximate Nearest Neighbor (ANN) model, is responsible for retrieving similar conversations from a database of embedding- and raw-text-tuples. Given the context of an unfinished conversation, suitable responses are considered to be contained in similar conversations, retrieved by the ANN model. The last model component receives a retrieved set and ranks possible answers based on answer- and context-relevance. The entire pipeline can be seen in Figure \ref{fig:pipelineoverview}.

\subsection{Gated Recurrent Unit (GRU)}
Recurrent Neural Networks (RNNs) have been designed to process sequential data by encoding historic information into a hidden state. When optimizing the network to predict future values using the current input and historic values, the hidden state naturally becomes a new representation form of the processed data. LSTMs, first introduced 1997, are an attempt to increase a RNN's capabilities to remember long-term dependencies by replacing hidden units with more complex memory cells, capable of controlling the information flow in and out of the cells.

The more recently proposed GRUs are similar to LSTM cells aiming to improve a RNN's capabilities to remember long-term dependencies. However, they use a different gate design, have fewer parameters to train and come without an additional cell state. Two gates, the reset and update gates $r_t$ and $z_t$, operate directly on the hidden state, i.e., the hidden layer. Parametrized by $W$, $U$ and $b$, while conditioned on the current input $x_t$ and previous result $y_{t-1}$, GRU gate vectors are computed as:
\begin{equation} 
\label{eq:gru1}
\begin{aligned}
      & z_t = \varphi_g(W_zx_t + U_zy_{t-1}+b_z) \\
      & r_t = \varphi_g(W_rx_t + U_ry_{t-1}+b_r),
\end{aligned}
\end{equation}

with $\varphi_g$ being the sigmoid function. The update gate $z_t$ combines the function of the input and forget gate by controlling how much the new hidden state ($h_t$) is defined by either the current input or the last hidden state, using linear interpolation:

\begin{equation}
\label{eq:gru2}
h_t = z_t \circ h_{t-1} + (1-z_t) \circ \tilde{h_t},
\end{equation}

with $\tilde{h_t}$ being the candidate activation. The reset gate $r_t$ is used to calculate $\tilde{h_t}$, controlling similarly how much of the previous hidden state to keep:

\begin{equation}
\label{eq:gru3}
\tilde{h_t} = \varphi_h(W_hx_t+ U_h(r_t \circ h_{t-1} + b_h),
\end{equation}

with $\varphi_h$ being the hyperbolic tangent function.

\subsection{Hierarchical Recurrent Encoder-Decoder}
The HRED model essentially consists of three stacked RNNs: the utterance encoder, context encoder, and utterance decoder, each of them depending on the result of its predecessor and operating on distributed representations. 

Formally, a dialogue $D$, the input to such a model, can be represented as a sequence of utterances $D = (U_1,...,U_M)$, with $U_m$ being a sequence of word indices  $U_m = (w_{m,1},...,w_{m,N_m})$, each of them usually pointing to a vocabulary reference or directly to a word embedding. These become the input to the utterance encoder.

\subsubsection{Encoding steps}

To better capture long-term dependencies, the GRU gating function is used for the individual RNNs. For a simplified notation, the GRU can be expressed compactly by combining equations \ref{eq:gru1}, \ref{eq:gru2} and \ref{eq:gru3}:

\begin{equation}
    h_t = GRU(h_{t-1}, x_t),
\end{equation}

computing current hidden state $h_t$, conditioned on the previous, $h_{t-1}$ and on current input $x_t$. To comply with the HRED notation, the utterance embedding $h_{m,n}$ of the current utterance $U_m$, including word $w_{m,n}$, is calculated as: 

\begin{equation}
\label{eq:hred1}
    h_{m,n} = GRU_{utt}(h_{m,n-1}, w_{m,n}).
\end{equation}

Applying equation \ref{eq:hred1} consecutively on word embeddings $w_{m,1},...,w_{m,N_m}$, results in an equally-sized set of hidden states $h_{m,1},...,h_{m,N_m}$, where the last hidden state $h_{m,N_m}$ is the summary of all words in the same utterance. As such, we denote $h_m=h_{m,N_m}$ to be the hidden state that represents utterance $U_m$.

Using this encoding approach, a set of utterances $U_1,...,U_M$ is encoded into hidden states $h_1,...,h_M$. Those are used as input to the GRU-based context encoder, similar to how word embeddings acted as input to the utterance encoder. As such, context embeddings $c_m$ are a summary of utterances and represent entire dialogues. They are computed as:

\begin{equation}
\label{eq:hred2}
    c_{m} = GRU_{con}(c_{m,n-1}, h_{m}).
\end{equation}

\subsubsection{Decoding step}

In addition to encoding a sequence of embeddings into a hidden state, the decoder component generates word probabilities over a vocabulary, given some context $U_{1},..., U_{m-1}$ and previous words $w_{m,1},..., w_{m,n-1}$.

Firstly, to condition the decoder RNN on previous utterances, the initialization of its hidden state is based on the context encoders last hidden state $c_{m-1}$. If not designed explicitly, context and decoder RNN usually have different hidden state dimensionalities, which is why an additional network layer is added to project context embeddings into the decoder space:

\begin{equation}
\label{eq:dec1}
    d_{m,0} = tanh(D_0c_{m-1}+b_0),
\end{equation}

with parameters $D_0$ and $b_0$ and $d_{m,0}$ being the decoder RNN's initial hidden state.

Given a set of words $w_{m,1},..., w_{m,n-1}$, having been previously generated or representing a training example, the decoder RNN hidden state is similarly computed as it was done for the encoder RNNs:

\begin{equation}
    d_{m,n} = GRU_{dec}(d_{m,n-1},w_{m,n}),
\end{equation}

processing words consecutively. The first iteration uses the hidden state computed by equation \ref{eq:dec1} and a zero-value embedding for $w_{m,0}$ to predict the first word of an utterance.

Using both, the hidden state $d_{m,n-1}$ and word embedding of $w_{m,n-1}$, the word embedding of current word $w_{m,n}$ is then predicted as:

\begin{equation}
    w(d_{m,n-1},w_{m,n-1})=H_0d_{m,n-1}+E_0w_{m,n-1}+b_0,
\end{equation}

with the additional parameters $H_0$, $E_0$ and $b_0$. $H_0$ and $E_0$ control which part of the previous context- and word embedding contribute to the new word embedding and how much of that part is used. 

By calculating the dot product of such generated word embeddings with the embeddings in a vocabulary, one can compute the similarity between prediction and existing words, with the most similar word being the most likely one. The actual probability of a word occurring next is based on this similarity:

\begin{equation}
\begin{split}
    P(w_{m,n} = v | w_{m,1:n-1}, U_{1:m-1}) = \\ 
    \frac{exp(e_v^\top w(d_{m,n-1},w_{m,n-1}))}{\sum_{k=1}^{K}exp(e_k^\top w(d_{m,n-1},w_{m,n-1}))},
\end{split}
\end{equation}

with $e_v$ being the word embedding of word $v$, $w$ the predicted word embedding and $K$ the vocabulary size.

By computing probabilities for each word in the vocabulary, one can create a distribution from which words can be sampled. Pushing sampled words back into the decoder allows to generate the next word, extending an utterance until an end-of-sequence meta token has been reached. Possible generations can be explored using probability based search techniques such as Beam Search \cite{cho2014learning}.

\subsection{Retrieval Model}
Using the encoded corpus as a database of vectors, a Nearest Neighbor Search (NNS) algorithm can be used to find close embeddings in the whole set. For this purpose, an ANN approach has been considered, as general space-partitioning approaches, aiming to improve the NNS performance, suffer from the curse of dimensionality \cite{weber1998quantitative}.

Locality Sensitive Hashing (LSH) \cite{har2012approximate, bawa2005lsh} is an ANN approach that uses a set of hashing functions to project similar data points into buckets and as such, significantly restricts the search space to the size of the bucket. For a projection, a binary string label is constructed by applying $k$ different hashing functions to a single data point, where the output of such a function is either one or zero. The desired goal of a hashing function is to output the same label for similar data points and differing labels for dissimilar ones. Therefore, binary string labels that are similar, indicate that also the original data points are similar. The string label is then used as a key to index a bucket of similar data points, where a brute force approach can be applied on a much smaller set. A collection of buckets is called a hash-table and $l$ tables constitute the entire model.

One of the main issues with the basic LSH algorithm \cite{bawa2005lsh} is that choosing the optimal number of hashing functions $k$ and number of tables $l$ requires one to know the most suitable value for $r$, the threshold separating similar and dissimilar points. The LSH-Forest algorithm solves this issue by allowing labels with variable length and thus, eliminating parameter $k$. 

Instead of linking fixed-length labels to buckets, the label string is stored in a prefix tree, a binary tree (also called 'trie'), in which keys are not contained within nodes but derived from the path that leads from the root to a node. 

For the case that two points are very similar or equal, the length-limiting parameter $k_m$ prevents their labels from growing too large.

Each level of the tree is associated with a different hashing function, sampled uniformly and with replacement from a family of hashing functions $\mathcal{H}$. Such a tree, an LSH-Tree, is the equivalent to an LSH-based hash-table and the composition of $l$ trees is an LSH-Forest.

Given a query point $p$, finding close neighbors in a set of LSH-Trees $T_1, T_2, ..., T_l$ is performed in two phases. First, in a top-down phase or descent, each tree $T_i$ is searched for the leaf node with the best match to the binary string label of $q$. Labels are computed individually for each tree, starting from the root and extending the label until a leaf node is reached.

Inspecting the matches from all trees, the match with the longest prefix defines the tree-level $x$ from which the bottom-up accumulation, the second phase, begins.

\subsection{Candidate Selection} \label{sec:candselection}

Given a query context of an unfinished conversation, using the previously discussed LSH-Forest algorithm, one can retrieve a candidate set from a database of encoded conversations.
Candidate answers will be scored based on the matching degree between the retrieved and the original context in terms of question-to-question similarity or answer relevance or other text-based features. However, the scoring functions introduced in this section will solely be based on vector comparison metrics, such as the cosine similarity, as text-based comparison is less rewarding and more difficult and tedious to implement. For the sake of clarity, the query context embedding is defined as $c_q$, the textual candidates as $r_1, r_2, ..., r_k$, the context embeddings of candidates as $c_{r_1}, c_{r_2}, ..., c_{r_k}$, and the utterance embeddings of candidates as $h_{r_1}, h_{r_2}, ..., h_{r_k}$.

\subsubsection{Context Relevance}

The similarity of two conversations or the distance between a query context $c_q$ and a candidate context $c_{r_k}$ has, intuitively, a big impact on the retrieved answer, i.e, the more two questions are similar, the higher the probability that the answers are similar as well. If the cosine similarity has been chosen as the distance function $D$, the labels returned by the nearest neighbor search are already sorted by context-to-context distance. Formally, given a candidate response $r_x$ and a query context $c_q$, the Context Relevance (CR) cost function is defined as:

\begin{equation}
    cost_{CR}(r_x) = cos_{sim}(c_{r_x}, c_{q}).
\end{equation}

\begin{figure}[h]
\centering
\scalebox{1}{
\includegraphics[width=0.35\textwidth]{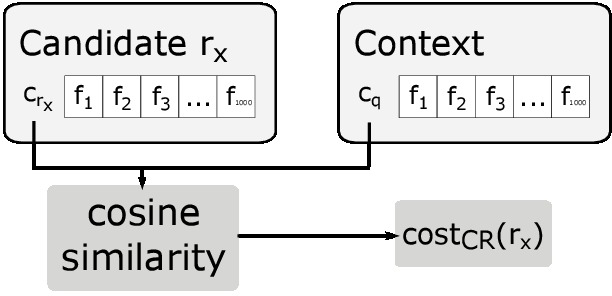}}
\caption{A simple illustration of how CR is computed for a single candidate.}
    \label{fig:AR}
\end{figure}

\subsubsection{Answer Relevance}

By manual inspection of near neighbors, it became apparent that the correct answer is usually represented or almost captured in many topic-related candidates. Assuming that the most suitable topic for answering is dominantly represented amongst candidates, responses are ranked based on how much they capture the general topic. Formally, the cost of a response $r_x$ is defined by the accumulated similarity between its respective embedding $h_{r_x}$ and the utterance embeddings of all other candidates (See Figure \ref{fig:AR}), normalized by length $k$:

\begin{equation}
    cost_{AR}(r_x) = \frac{1}{k}\sum_{i=1}^{k} cos_{sim}(h_{r_x}, h_{r_i})
\end{equation}

\begin{figure}[h]
\centering
\scalebox{1}{
\includegraphics[width=0.35\textwidth]{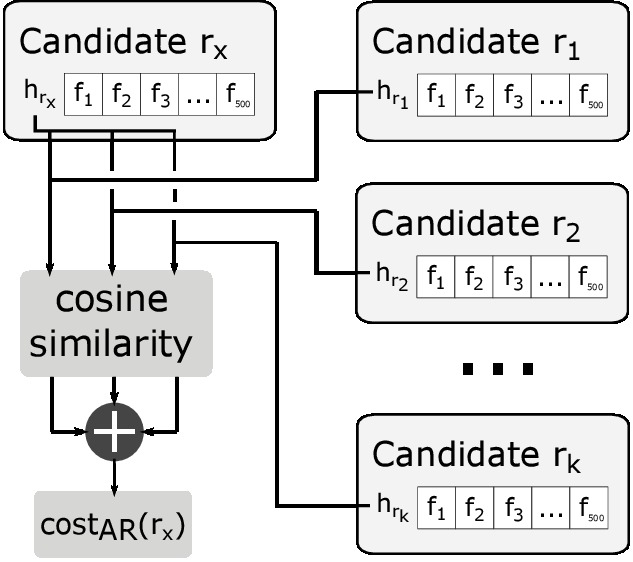}}
\caption{An image showing how the AR cost of a single candidate is accumulated by computing the cosine similarity with other candidate's utterance embeddings.}
    \label{fig:AR}
\end{figure}

\subsubsection{Combining Context and Answer Relevance}

The problem with the previous approach is that the candidates that are off-topic still contribute to the answer relevance cost. Therefore, in a pre-step, according to the previously described context relevance metric, the top $n$ candidates are accumulated to represent the best general answer topic. In the next step, candidates are ranked based on their similarity to these $n$ responses. Formally, combined Context and Answer Relevance (CAR) is defined as:
\begin{equation}
    cost_{CAR}(r_x) = \frac{1}{n}\sum_{i=1}^{n} cos_{sim}(h_{r_x}, h_{r_i}),
\end{equation}
with $n \leq k$.

\section{Experiments} \label{experiments}

\subsection{Datasets} \label{corpus}
The first dataset we use is the Ubuntu Dialogue Corpus which has been studied in most state of the art systems (similar to HRED). The ubuntu dataset contains almost 1 million multi-turn dialogues, with a total of over 7 million utterances and 100 million words. More information can be found in \cite{lowe2015ubuntu}.

The second dataset we use is the Vodafone corpus which is created by retrieving archived conversations of the Dutch Vodafone online customer service. Customers having problems with their phone, want to make contractual changes or experience other product related issues, often decide to talk with a Vodafone service agent through an online chat platform. 


Every conversation that was not clearly identified as Dutch text was filtered out of the corpus using a port of Google's Java language detection implemention \footnote{http://code.google.com/p/language-detection/}. Furthermore, to guarantee that the HRED model receives actual conversations for its training, conversations that have less than $5$ turns have also been filtered out.

The original corpus contains phone numbers, addresses, names, postal codes and other personal information. To guarantee anonymization and also to allow enough generalization, this data has been replaced by a meta-token, e.g. \texttt{"<street\_name>"} or \texttt{"<city>"}, which is considered to be beneficial for the performance of word embeddings. This way many more training examples will contain these general concepts (like \texttt{"<street\_name>"}) and the model can learn in which context a street name should appear. This is possible because the word embeddings of such concepts are also tuned during the training.

The final corpus was generated by using a minimal word occurrence threshold of $10$, resulting in a dictionary size of $42892$ and an average of $0.435\%$ unknowns per dialogue. The complete statistics for both datasets can be seen in Table \ref{table:statsfinal}.

\begin{table}[!htb]
\begin{center}
\scalebox{1}{
\begin{tabular}{ |c|c|c| } 
 \hline
 & Ubuntu & Vodafone \\
 \hline
 Language & English & Dutch\\
 Total \# of dialogues & 487,337 &384,897  \\ 
 Total \# of turns & 2,406,483 & 6,571,902  \\ 
 Total \# of utterances & 3,644,566 & 10,461,677  \\ 
 Total \# of words & 44,246,198 &122,325,433  \\ 
 Avg. \# of words per dialogue & 90.792 & 317.81  \\ 
 Avg. \# of turns per dialogue & 4.938 & 17.07  \\ 
 Avg. \# of words per turn &15.880 & 18.65  \\ 
 Avg. \# of utterances per dialogue &  7.479 & 27.18  \\ 
 Avg. \# of words per utterance & 11.264 & 11.58  \\
 \hline
\end{tabular}}
\end{center}
\caption[Final statistics of the used corpus]{Statistics of Ubuntu \& Vodafone Corpus}
\label{table:statsfinal}
\vspace{-20pt}
\end{table}

\subsection{Evaluation process} \label{evaluation_metric}
A quantitative evaluation metric, the Recall@k measurement \cite{lowe2015ubuntu}, has been used to compare the ranking performance of models. Given a context, a set of $n$ possible answers is presented to a model, which has to rank the answers by their likelihood of being the actual response. For a single evaluation sample, if the correct answer is ranked to be amongst the $k$ best, the model succeeded. The overall performance of a model is defined as the ratio of correctly ranked answers to all answers, i.e., the percentage of correct answers that were ranked to be amongst the $k$ best.

By iterating over the conversations in the held-out test set, an evaluation sample has been created for each individual turn or response, with the previous turns representing the context and the current turn or response being the ground truth. In addition to the actual response, a single example also contains $n-1$ randomly sampled answers, which the model should preferably rank lower than the true answer. 



Each of the models, generative- or retrieval-based, receives the context of a conversation from an evaluation sample and has to generate or retrieve a suitable answer. The utterance-embedding of this answer is then used to compute the distance to each of the $10$ possible answers in the evaluation sample, using the cosine similarity between utterance-embeddings. The final ranking is based on this distance, placing similar answers at the top. 

As we wanted to have the same conditions for all models, our ranking approach differs from the one used in \cite{lowe2015ubuntu}, where answer-embeddings have been directly predicted by an additional network layer. Instead, we used Beam Search (using 5 beams) to generate an answer with the $HRED$ model and used the answer's utterance embedding to compute the ranking for the generative approach. 

\subsection{Results and discussion} \label{results_and_discussion}
For the Ubuntu corpus, a HRED model is trained and then the generative approach (of the original model) and the different candidate selection methods (as described in Section \ref{sec:candselection}) are compared. Results can be found in Table \ref{table:ubuntu_results}.

\begin{table}[h]

\centering
\scalebox{1.0}{
\begin{tabular}{l|l|l|l|}
\hline
\multicolumn{1}{|l|}{Model}            & R@1                           & R@2                           & R@5                          \\ \hline
\hline
\multicolumn{1}{|l|}{$HRED$}                & $34.8\pm0.4$ & $50.5\pm0.4$ & $78.2\pm0.3$
 \\ \hline \hline
\multicolumn{1}{|l|}{$HRED$-$CR$}                & $32.8\pm0.3$ & $47.5\pm0.4$ & $74.1\pm0.3$
\\\hline
\multicolumn{1}{|l|}{$HRED$-$AR$}                & $\mathbf{44.1\pm0.4}$ & $\mathbf{58.6\pm0.4}$ & $\mathbf{80.5\pm0.3}$
\\\hline
\multicolumn{1}{|l|}{$HRED$-$CAR$}                & $43.5\pm0.4$ & $58.0\pm0.4$ & $80.3\pm0.3$   
\\\hline
\end{tabular}
}

\caption[Overall ranking performance on the ubuntu corpus between all models]{Overall ranking performance of models on the ubuntu corpus. Confidence intervals ($\pm 95\%$) are shown next to the average performance. }

\label{table:ubuntu_results}
\end{table}

From this Table it is obvious that AR model outperforms other candidate selection techniques as well as the generative approach. When taking into account the context of the whole conversation, results are slightly worse which means that answers are better predicted by focusing on each turn individually rather than taking into account the entire context.

For the Vodafone corpus, three HRED models have been compared, each initialized with a different set of word embeddings in order to assess the effect of local/global context in a language setting other than English. The first model, based only on local domain knowledge, received word embeddings trained with the gensim python library \cite{rehureklrec}, a tool that, given a corpus, will train word embeddings specifically for that corpus. The second model utilized word embeddings from \cite{tulkens2016evaluating}, representing global domain knowledge. The embeddings were trained on a corpus consisting of $4$ billion words, which was automatically generated by analyzing Dutch websites. The last model received word embeddings that are a combination of the two previously described sets. Both contain word embeddings with a feature-length of $320$. However, the embeddings for the last model will have a length of $420$, using a concatenation of global embeddings with $320$ features and local embeddings with $100$ features.

As with the Ubuntu corpus, the generative and retrieval based approaches ($CR$, $AR$, $CAR$) are compared and additionally in this setting, they are also tested upon different embedding initialization approaches ($HRED_L$, $HRED_G$ and $HRED_{LG}$) An overview of the results can be found in Table \ref{table:overal}. 

\begin{table*}[h]
\centering
\scalebox{1}{
\begin{tabular}{ll|l|l|l||l|l|l|}
\cline{3-8}

                                  &     & \multicolumn{3}{l||}{\begin{tabular}[c]{@{}l@{}}Predicting assistant\\ responses\end{tabular}} & \multicolumn{3}{l|}{\begin{tabular}[c]{@{}l@{}}Predicting customer\\ responses\end{tabular}} \\ \hline
\multicolumn{1}{|l|}{Approach} & {Model}            & R@1                           & R@2                           & R@5                          & R@1                           & R@2                           & R@5                           \\ \hline
\hline

\multicolumn{1}{|l|}{\multirow{3}{*}{Beam Search}}    & $HRED_L$           & $\mathbf{31.2\pm0.9}$ & $\mathbf{45.8\pm0.8}$ & $\mathbf{73.4\pm0.9}$                       & $28.5\pm0.7$ & $\mathbf{39.8\pm0.9}$ & $\mathbf{66.1\pm0.8}$                                     \\ \cline{2-8}

\multicolumn{1}{|l|}{} & $HRED_G$             & $29.9\pm0.7$ & $44.3\pm0.9$ & $71.1\pm0.6$                              & $27.1\pm0.7$ & $37.7\pm0.9$ & $63.1\pm0.9$                        \\ \cline{2-8}

\multicolumn{1}{|l|}{} & $HRED_{LG}$             & $30.3\pm1.0$ & $44.2\pm0.9$ & $70.6\pm0.7$                            & $\mathbf{28.9\pm1.1}$ & $\mathbf{39.8\pm1.1}$ & $64.6\pm1.2$                    \\ \hline

\hline

\multicolumn{1}{|l|}{\multirow{3}{*}{Context Relevance}} & $HRED_{L}$-$CR$     & $33.4\pm0.7$ & $48.0\pm0.8$ & $75.4\pm0.6$                           & $32.8\pm1.0$ & $47.5\pm1.0$ & $73.5\pm0.9$             \\ \cline{2-8}

\multicolumn{1}{|l|}{} & $HRED_{G}$-$CR$       & $34.6\pm1.1$ & $50.0\pm1.0$ & $76.2\pm0.7$        & $32.9\pm0.8$ & $47.3\pm0.7$ & $74.2\pm0.8$                            \\ \cline{2-8}

\multicolumn{1}{|l|}{} & $HRED_{LG}$-$CR$  & $33.9\pm0.9$ & $48.8\pm0.8$ & $74.8\pm0.8$                           & $32.5\pm0.9$ & $48.1\pm0.7$ & $74.8\pm0.6$                               \\ \hline

\multicolumn{1}{|l|}{\multirow{3}{*}{Answer Relevance}} & $HRED_{L}$-$AR$       & $39.6\pm0.7$ & $55.7\pm0.9$ & $81.1\pm0.7$                             & $40.0\pm1.1$ & $55.8\pm1.0$ & $80.4\pm0.8$                     \\ \cline{2-8}

\multicolumn{1}{|l|}{} & $HRED_{G}$-$AR$      & $42.7\pm0.9$ & $58.5\pm0.8$ & $82.5\pm0.7$                            & $\mathbf{41.0\pm0.9}$ & $\mathbf{56.9\pm0.7}$ & $\mathbf{81.5\pm0.6}$                            \\ \cline{2-8}

\multicolumn{1}{|l|}{} & $HRED_{LG}$-$AR$  & $43.0\pm0.8$ & $59.4\pm0.8$ & $\mathbf{82.7}\pm0.8$                        & $40.1\pm0.9$ & $55.7\pm0.9$ & $80.0\pm0.9$                              \\ \hline

\multicolumn{1}{|l|}{\multirow{3}{*}{\shortstack{Context and Answer\\ Relevance}}} & $HRED_{L}$-$CAR$      & $41.3\pm0.8$ & $57.2\pm0.8$ & $81.2\pm0.5$                            & $39.9\pm1.0$ & $55.3\pm1.0$ & $79.6\pm0.6$                             \\ \cline{2-8}

\multicolumn{1}{|l|}{} &$HRED_{G}$-$CAR$      & $\mathbf{44.0\pm0.7}$ & $\mathbf{59.8\pm0.9}$ & $82.6\pm0.7$                         & $40.9\pm0.7$ & $56.8\pm0.7$ & $80.6\pm0.7$                                \\ \cline{2-8}

\multicolumn{1}{|l|}{} &$HRED_{LG}$-$CAR$ & $43.8\pm0.7$ & $59.5\pm0.8$ & $82.6\pm0.7$                            & $39.4\pm0.7$ & $55.1\pm0.9$ & $79.6\pm0.6$                        \\ \hline

\end{tabular}
}
\caption[Overall performance comparison between all models]{Overall comparison of model precisions (in $\%$). Confidence intervals ($\pm 95\%$) are shown next to the average performance. }
\label{table:overal}
   \vspace{-20pt}
\end{table*}

As expected, for the majority of setups, models can predict assistant responses easier than customer responses. The CAR candidate selection method outperforms all other techniques when predicting assistant responses. However, the performance of customer response prediction is slightly dominated by AR. A reason for this could be that customers often reply with new questions that might not be context related, making answer relevance more important than context relevance.

Furthermore, it can be seen that initializing the HRED model with word embeddings containing global domain knowledge results in the best performance for candidate selection approaches. However, combining global and local domain knowledge has not led to the desired improvements. This can be explained as follows: The computational graph of the HRED model that defines its training also includes tuning the word embeddings. As such, during the training, the word embeddings are already altered to encode local domain knowledge, even if they were only initialized with embeddings containing global domain knowledge. Adding additional feature-length will in the worst case only add complexity to the model.

The performance of the generative approaches, $HRED_L$, $HRED_G$, and $HRED_{LG}$, are relatively similar. However, $HRED_L$ slightly outperforms the others. This is contradictory, considering that the candidate selection methods, CR, AR and CAR, clearly perform better on embeddings generated by $HRED_G$ and $HRED_{LG}$ (See Table \ref{table:overal}). A reason for this could be that utilizing global domain knowledge to generate an answer is more difficult than using specific domain knowledge. Especially for a very homogeneous (and domain specific) corpus, giving standard answers can work better. Nonetheless, similarity comparisons, used by the NNS approaches, could still benefit from richer embeddings.  

Table \ref{table:ubuntu} presents some examples of answers using the Ubuntu Corpus using the generative approach (HRED) and the proposed AR model. Finally, in Table \ref{table:chatgood} 
we present one chat example from the evaluation process on Vodafone corpus (translated in English by the authors).

\begin{table*}[h]
\centering
\caption{Answers of the HRED model and the retrieval-based approach (using AR) to a couple of questions from the Ubuntu corpus test set. Questions considered interesting and clean where picked by the author.}
\label{table:ubuntu}
\scalebox{0.9}{
\begin{tabular}{|l|l|}
\hline
Context                                                                                                                                                                                                                                                                                                                                            & Response                                                                                                                                                                                                                                                                              \\ \hline \hline
\begin{tabular}[c]{@{}l@{}}I have a netbook \_\_eou\_\_ with 1gb of ram . Will ubuntu 12.04 be fine ? \\ \_\_eou\_\_ With unity of course !\end{tabular}                                                                                                                                                                                                                                                                                                                                                                & \begin{tabular}[c]{@{}l@{}}\textbf{Original}: barely\\ \textbf{AR}: you don't need much hardware to run gnome , \\ unity perhaps yes though\\ \textbf{HRED}: that 's a good idea , \\ but it 's not a good idea to stick with it\end{tabular}                                                                    \\ \hline
\begin{tabular}[c]{@{}l@{}}Hi Guys \_\_eou\_\_ I need urgent help \_\_eou\_\_ I've been trying all day to \\ install Ubuntu on my Mac \_\_eou\_\_ yeah but I can't find ANY solution \\ \_\_eou\_\_ but can someon **unknown** me determining what to do ? \\ \_\_eou\_\_ My **unknown** USB gives me kernel panic \_\_eou\_\_ \\ not syncing \\ \\ \end{tabular}                                                                                                                                                              & \begin{tabular}[c]{@{}l@{}}\textbf{Original}: Why dont you use a cdrom for \\ installing Ubuntu ?\\ \textbf{AR}: does your Bios support booting usb device ?\\ \textbf{HRED}: not sure what to tell you , sorry\end{tabular}                                                                                     \\ \hline
\begin{tabular}[c]{@{}l@{}}any help folks **unknown** \_\_eou\_\_ $\rightarrow$ help with what ? \_\_eou\_\_ $\rightarrow$\\  \textless **unknown** \textgreater hello folks , ok so i will try this again maybe \\ someone here is able to help me with my issue . running 12.04 on an hp \\ **unknown** laptop . when i plug my headphones sound still comes \\ from main speakers and my headphone , how can i make it so that when \\ i plug my headphone the sound will only play thru the headphones \\ \\ \end{tabular} & \begin{tabular}[c]{@{}l@{}}\textbf{Original}: did you check out the **unknown** \\ link someone posted to you ?\\ \textbf{AR}: make sure you have the mixer and \\ sound devices installed in your kernel\\ \textbf{HRED}: that sounds like a hardware issue , right ?\end{tabular}                              \\ \hline
\begin{tabular}[c]{@{}l@{}}Why can't I install Ubuntu ? I burned the ISO to a disk , ran the disk on \\ windows , it **unknown** installed , but all I can get is the demo version . \\ And when I try to directly install the FULL version , it doesn't do anything \\ but restart my computer . And the **unknown** won't even work\\ unless I boot from CD . \\ \\ \end{tabular}                                                                                                                                            & \begin{tabular}[c]{@{}l@{}}\textbf{Original}: what is the ISO named ? **unknown** - \\ are you sure that 's " ubuntu "?\\ \textbf{AR}: have you simply tried to burn a new cd ?\\ \textbf{HRED}: you have to use the alternate cd to install \\ the alternate cd\end{tabular}                                    \\ \hline
my computer crashed ? how can I see the log ?                                                                                                                                                                                                                                                                                                                                                                                                                                                                           & \begin{tabular}[c]{@{}l@{}}\textbf{Original}: look in /var/log - especially at the xorg log \\ \textbf{AR}: somewhere in /var/log\\ \textbf{HRED}: it should be in the menu . lst file\end{tabular}                                                                                                              \\ \hline
\begin{tabular}[c]{@{}l@{}}Question : is sshd installed by default with ubuntu 12.04 or do i need\\ to install openssh\end{tabular}                                                                                                                                                                                                                                                                                                                                                                                     & \begin{tabular}[c]{@{}l@{}}\textbf{Original}: normally installed by default\\ \textbf{AR}: ubuntu doesn't install an ssh daemon by default \\ so yes\\ \textbf{HRED}: it should be installed by default\end{tabular}                                                                                             \\ \hline
\begin{tabular}[c]{@{}l@{}}is there a " terminal service " for linux \_\_eou\_\_ i.e. not vnc \\ \_\_eou\_\_ $\rightarrow$  X ? \_\_eou\_\_ ssh ? \_\_eou\_\_\end{tabular}                                                                                                                                                                                                                                                                                                                                              & \begin{tabular}[c]{@{}l@{}}\textbf{Original}: i tried it and it was rather laggy\\ \textbf{AR}: idk it just seems lame lol ; x\\ \textbf{HRED}: this is what I was looking for\end{tabular}                                                                                                                      \\ \hline
How hard will it be to upgrade from Ubuntu 12.04 to 12.10 ?                                                                                                                                                                                                                                                                                                                                                                                                                                                             & \begin{tabular}[c]{@{}l@{}}\textbf{Original}: update-manager will prompt you if you want to \\ upgrade when it 's released and you press yes\\ \textbf{AR}: it is still there , just not installed by default\\ \textbf{HRED}: first you have to use sudo apt-get dist-upgrade \\ to get it to work\end{tabular} \\ \hline
\end{tabular}}
\end{table*}
\begin{table}[h]

\centering
\resizebox{\columnwidth}{!}{
\begin{tabular}{|l|}
\hline
\textbf{Context} \\ \hline
\begin{tabular}[c]{@{}l@{}} \texttt{<customer>} my phone is damaged, \\ it is new, do i get a new phone? \texttt{</u>} \\ \texttt{<assistant>} did the phone fall? \texttt{</u>} \\ \texttt{<customer>} no \texttt{</u>} \\ \texttt{<assistant>}  when did you got the phone? \texttt{</u>} \\ \texttt{<customer>} I had the phone in my pocket, just when taking it out \\ I saw that maybe it was scratched by my keys \texttt{</u>} \texttt{<month>} \texttt{</u>} \\ \\ \end{tabular} \\ \hline 
\textbf{Actual Response} \\ \hline
\texttt{<assistant>} have you got an insurance? \texttt{</u>} \\ \\ \hline
\textbf{$\mathbf{HRED_G}$} \\ \hline
\texttt{<assistant>} I will gladly check it for you \texttt{</u>} \\ does it concern \texttt{<number>} ? \texttt{</u>} \\ \\ \hline
$\mathbf{HRED_G}$-$\mathbf{CAR}$ Candidates (best three) \\ \hline
\begin{tabular}[c]{@{}l@{}} \texttt{<assistant>} you have \texttt{<number>} years warranty for your device. \\ internal device issues are covered.\\  if you haven't damaged the phone yourself, \\ it falls under the regular warranty. \texttt{</u>} \\ do you have an insurance with us? then you are insured either way. \texttt{</u>} \\ \\ \texttt{<assistant>} here they actually can check your device \\ and if necessary, send it in for repair. \texttt{</u>} \\ oh ... you have an insurance? \texttt{</u>} \\ \\ \texttt{<assistant>} You can then send your device in for repair. \texttt{</u>} \\ if you have the device replacement service in your subscription \\ your device will be \texttt{<unk>} and you will receive a replacement device. \\ In case you don't have it? \texttt{</u>} \\ then you can bring your device \texttt{</u>} to a Vodafone shop. \texttt{</u>}\end{tabular} \\ \\ \hline
\end{tabular}}

\caption[Chat example using the proposed model]{An example showing the three best responses retrieved by the $HRED_G$-$CAR$ model to a given context. For comparison reasons, the answer generated by the $HRED_G$ model and the actual response are shown as well.}
\label{table:chatgood}

\end{table}

\section{Conclusion and Future Work} \label{conclusion}

End-to-End Dialogue Systems are relatively new and most architectures are far away from being ready for deployment in actual industry that most likely will require more years of research. The architecture proposed in this paper can be seen as a combination of end-to-end and modular Dialogue System. The used retrieval based approach, utilizing dialogue and utterance embeddings which were trained end-to-end, has been shown to outperform the generative approach of the HRED model. 

More recently proposed end-to-end systems, the VHRED model and Multiresolution Recurrent Neural Networks \cite{serban2016multiresolution}, both being an improved version of the HRED model, are raising another question: Will one of these models outperform the proposed retrieval based approach, even though all operate on the same embeddings, i.e., at which point does the generative approach benefit from the embeddings' quality more than the retrieval-based? Dialogue and utterance embeddings have been only generated by one source, the HRED model. For comparison reasons, it would be interesting to explore the performance of other encoding approaches, such as averaging over word embeddings. Additionally, the embeddings generated by the recently proposed VHRED model and Multiresolution Recurrent Neural Networks are expected to be of higher quality and likely to improve the performance of the proposed approach. However, one must not underestimate the importance of the proposed ranking system, since it can directly be used by a human agent as a means to assist communication with a client. An interesting research direction arising from this paper would be to allow human agents to affect the ranking score and by this way providing feedback (in terms of reinforcement learning \cite{williams2016end}) to the system, which then might be able to re-rank answers.

Another future direction is the simulation of conversations with a tree search, by representing context embeddings as states (tree nodes) and utterance embeddings as actions (connections between nodes). This technique is inspired by game simulations where the desirable state is found through an exploration/exploitation strategy. Using responses retrieved by NNS as a set of actions, the search tree can explore possible paths and score responses based on the quality of simulated conversations.

Moreover, the candidate selection module of our proposed pipeline reveals new opportunities for utilizing such a ranking/similarity model in other problems/domains such as recommender systems. Reviews of products, services, etc. could be encoded using a model like HRED and then based on a query (question) of a user, recommendation can take place by ranking the most relevant reviews (answers).

Finally, another research direction towards the implementation of such a dialogue system, is the utilization of additional context information a service agent can see and that is not contained in the conversation. Considering the size of corpora used to train end-to-end systems (usually around $500,000$ conversations), manual annotation can be very slow and costly. Finding an automated approach to make addresses, contractual details and other features accessible to an end-to-end Dialogue System is an interesting and rewarding task. Solving (some of) the aforementioned problems will facilitate the deployment of end-to-end Dialogue Systems in online chat service environments, improving robustness, utility and customer experience.

\section*{Acknowledgement}
We would like to thank Harry Beckers and Marcel Overdijk for their collaboration and support. We gratefully acknowledge the support of QNH Consulting with the donation of the Nvidia GTX 1070 GPU used for this research and for providing the Vodafone dataset.

\bibliographystyle{IEEEtran}
\bibliography{IEEEabrv,references}

\end{document}